\documentclass[review]{elsarticle}
\graphicspath{ {./figures/} }

\usepackage{adjustbox}
\usepackage{hhline}
\usepackage{svg}
\usepackage{caption}
\usepackage{afterpage}
\usepackage{lipsum}

\usepackage{subcaption}
\usepackage{siunitx}

\usepackage{pifont}

\usepackage{verbatim}
\usepackage{apalike}

\usepackage{ragged2e}
\usepackage[normalem]{ulem}
\usepackage{amsthm}
\usepackage[mathscr]{euscript}
\usepackage{algorithmicx}
\usepackage{algpseudocode}
\usepackage{stackengine}
\usepackage[linesnumbered,ruled,vlined]{algorithm2e}
\newcolumntype{P}[1]{>{\centering\arraybackslash}p{#1}}
\newcolumntype{M}[1]{>{\centering\arraybackslash}m{#1}}
\usepackage[utf8]{inputenc}
\usepackage[fleqn]{amsmath}
\usepackage{bm}

\usepackage{mathtools}

\DeclareFontFamily{OT1}{pzc}{}
\DeclareFontShape{OT1}{pzc}{m}{it}{<-> s * [0.900] pzcmi7t}{}
\DeclareMathAlphabet{\mathpzc}{OT1}{pzc}{m}{it}
\usepackage{threeparttable}
\usepackage{booktabs}
\usepackage{soul}
\usepackage{pgfplots}

\usepackage{footmisc}
\usepackage{scrextend}
\usepackage{graphicx}
\usepackage{amsfonts}
\usepackage{enumitem}
\usepackage[mathscr]{euscript}
\usepackage{mathtools}
\usepackage{amssymb}
\usepackage{array}
\usepackage{float}
\usepackage{threeparttable}
\usepackage{amsmath}
\usepackage{blindtext}
\usepackage[utf8]{inputenc}
\usepackage{multirow}
\usepackage{color}
\usepackage{setspace}
\usepackage{hyperref}
\hypersetup{pdfauthor={Name}}

\usepackage{xcolor,colortbl}

\definecolor{gaincolor}{RGB}{230, 245, 255}

\usepackage{booktabs, makecell, siunitx, xcolor}
\usepackage[left=2.5cm,right=2.5cm,top=2cm,bottom=1.5cm]{geometry}

\journal{Computers in Biology and Medicine}

\biboptions{authoryear}

\begin{document}
\begin{frontmatter}

\begin{titlepage}
\begin{center}
\vspace*{1cm}

\textbf{\large Cytoplasmic Strings Analysis in Human Embryo Time-Lapse Videos using Deep Learning Framework}

\vspace{1.5cm}

Anabia Sohail$^{1,*}$ (anabia.sohail@ku.ac.ae), Mohamad Alansari$^{1,*}$ (100061914@ku.ac.ae), Ahmed Abughali$^{2,3}$ (100061920@ku.ac.ae), Asmaa Chehab$^{1}$ (100061201@ku.ac.ae), Abdelfatah Ahmed$^{1}$ (100059689@ku.ac.ae), Divya Velayudhan$^{1}$ (divya.velayudhan@ku.ac.ae), Sajid Javed$^{1,4}$ (sajid.javed@ku.ac.ae), Hasan Al Marzouqi$^{1}$ (hasan.almarzouqi@ku.ac.ae), Ameena Saad Al-Sumaiti$^{2,3}$ (ameena.alsumaiti@ku.ac.ae), Junaid Kashir$^{5}$ (junaid.kashir@ku.ac.ae), and Naoufel Werghi$^{1,4,6}$ (naoufel.werghi@ku.ac.ae) \\

\hspace{10pt}

\begin{flushleft}
\small  
$^{1}$Department of Computer Science, Khalifa University, Abu Dhabi, United Arab Emirates. \\
$^{2}$Department of Electrical Engineering, Khalifa University, Abu Dhabi, United Arab Emirates. \\
$^{3}$Smart OR Lab, Advanced Power and Energy Research Center, Khalifa University, Abu Dhabi, United Arab Emirates. \\
$^{4}$Center for Autonomous Robotic Systems, Khalifa University, Abu Dhabi, United Arab Emirates. \\
$^{5}$Department of Biology, College of Arts and Sciences, Khalifa University, Abu Dhabi, United Arab Emirates. \\
$^{6}$Center for Cyber-Physical Systems (C2PS), Khalifa University, Abu Dhabi, United Arab Emirates. \\
$^{*}$Indicates Equal Contributions. \\

\vspace{1cm}
\textbf{Corresponding Author:} \\
Mohamad Alansari \\
Department of Computer Science, Khalifa University, Abu Dhabi, United Arab Emirates.\\
Email: 100061914@ku.ac.ae

\end{flushleft}        
\end{center}
\end{titlepage}

\begin{abstract}
\noindent Infertility is a major global health issue, and while in-vitro fertilization (IVF) has improved treatment outcomes, embryo selection remains a critical bottleneck. Time-lapse imaging (TLI) enables continuous, non-invasive monitoring of embryo development, yet most automated assessment methods rely solely on conventional morphokinetic features and overlook emerging biomarkers. Cytoplasmic Strings (CS), thin filamentous structures connecting the inner cell mass and trophectoderm in expanded blastocysts, have been associated with faster blastocyst formation, higher blastocyst grades, and improved viability. 
However, CS assessment currently depends on manual visual inspection, which is labor-intensive, subjective, and severely affected by detection and subtle visual appearance. In this work, we present, to the best of our knowledge, the first computational framework for CS analysis in human IVF embryos. We first design a human-in-the-loop annotation pipeline to curate a biologically validated CS dataset from TLI videos, comprising 13,568 frames with highly sparse CS-positive instances. Building on this dataset, we propose a two-stage deep learning framework that (i) classifies CS presence at the frame level and (ii) localizes CS regions in positive cases. To address severe imbalance and feature uncertainty, we introduce the Novel Uncertainty-aware Contractive Embedding (NUCE) loss, which couples confidence-aware reweighting with an embedding contraction term to form compact, well-separated class clusters. NUCE consistently improves F1-score across five transformer backbones, while RF-DETR-based localization achieves state-of-the-art (SOTA) detection performance for thin, low-contrast CS structures. The proposed dataset, loss function, and detection pipeline establish a foundation for integrating CS as a robust biomarker in automated embryo assessment.
The source code will be made publicly available at: \url{https://github.com/HamadYA/CS_Detection}.
\end{abstract}


\end{frontmatter}

\newpage

\section{Introduction} \label{intro}
\noindent Infertility represents a major global reproductive health concern with substantial medical, psychological, and social implications, often diminishing individuals’ overall quality of life \cite{ahm0}. In vitro fertilization (IVF), an assisted reproductive technology, has emerged as an effective approach for managing infertility. Despite its transformative potential, IVF success rates remain modest (30\% to 50\%) with roughly one-third of cycles achieving clinical pregnancy and fewer resulting in live birth \cite{ahm1}.


\begin{figure}[t!]
\centering
\includegraphics[width=0.8\linewidth]{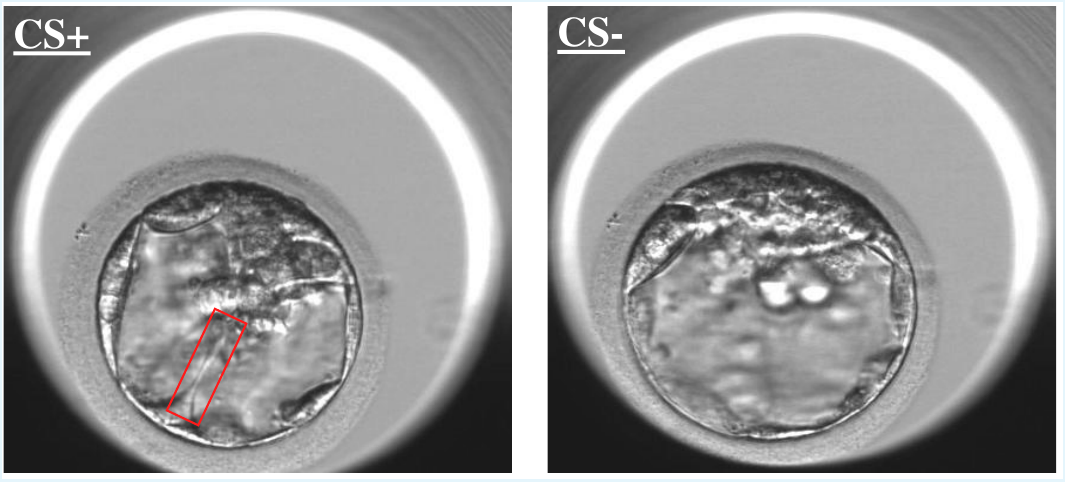}
\caption{Expanded blastocyst with CS (CS+ group) and without CS (CS- group). One CS is shown in the CS+ embryo traversing the blastocoel cavity (red arrows) and maintaining a connection between the ICM and the mural TE cells.}
\label{fig:1}
\end{figure}

\begin{figure}[t]
  \centering
  \begin{tikzpicture}
    \begin{axis}[
      ybar,
      width=0.7\linewidth, height=4.2cm,
      bar width=10pt,
      ymin=68, ymax=94.5,
      ylabel={F1-Score (\%)},
      symbolic x coords={ViT-B, Swin-B, CLIP},
      xtick=data,
      legend style={at={(0.5,1.05)},anchor=south,legend columns=-1},
      ymajorgrids,
      enlarge x limits=0.2,
      nodes near coords,
      nodes near coords align={vertical},
      nodes near coords style={font=\footnotesize, /pgf/number format/precision=2}
    ]
      \addplot coordinates {(ViT-B,82.71) (Swin-B,81.15) (CLIP,70.92)};
      \addplot coordinates {(ViT-B,88.67) (Swin-B,89.00) (CLIP,90.87)};
      \legend{Cross Entropy,NUCE}
    \end{axis}
  \end{tikzpicture}
  \caption{Performance comparison of baseline architectures using Cross-Entropy and the proposed NUCE loss. The \textbf{N}ovel \textbf{U}ncertainty-aware \textbf{C}ontractive \textbf{E}mbedding (NUCE) objective consistently improves F1-score across all backbone models (ViT-B, Swin-B, and CLIP).}
  \label{fig:bar}
  \vspace{-1em}
\end{figure}
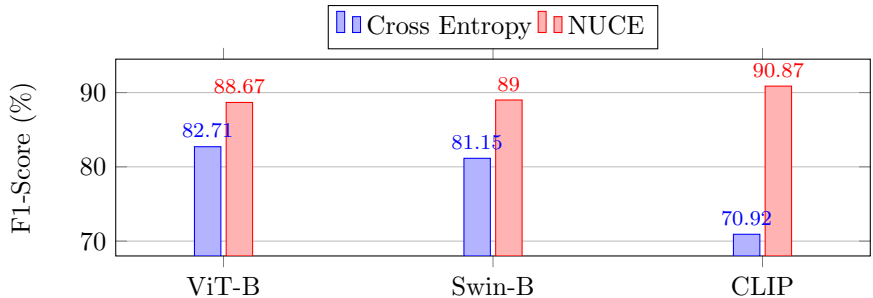

A critical determinant of IVF success is the accurate selection of the most viable embryo at the appropriate time. This has gained increasing importance with the widespread adoption of single-embryo transfer protocols, which aim to minimize the risks associated with multiple gestations \cite{ahmm}. Traditionally, embryologists have assessed embryo quality at discrete, static time points \cite{[1]}, an approach that provides limited insight into the embryo’s dynamic developmental potential.

The introduction of time-lapse imaging (TLI) has enabled continuous, non-invasive monitoring, generating comprehensive morphokinetic datasets \cite{[3], ahm6}. This technology allows embryologists to utilize dynamic features, such as cleavage timing and morphological changes, to identify embryos with the highest likelihood of implantation. However, most existing assessment systems primarily rely on standard morphokinetic parameters while overlooking additional biomarkers.

One such promising biomarker is the cytoplasmic string (CS). These are thin, filamentous structures that span the blastocoel, connecting the inner cell mass (ICM) to the trophectoderm (TE), as illustrated in Fig. \ref{fig:1}. 
Empirical evidence suggests that embryos exhibiting a higher number of CS, along with active vesicular transport along these structures, tend to achieve faster blastocyst formation and higher blastocyst grades \cite{[40]}.  
It is hypothesized that CS mediates intercellular communication between the ICM and TE during blastocoel expansion, a role further supported by the presence of key developmental receptors on these structures, such as growth factor 2 (FGF2) receptors and Erb-B2 receptor tyrosine kinase 3 (ErbB3) \cite{[41]}.

Despite their biological significance, the identification and analysis of CS have thus far relied exclusively on manual observation of time-lapse microscopy videos. This process is labor-intensive, expertise-dependent, and prone to inter- and intra-observer variability \cite{ahm5}. Moreover, the CS appears only in the later stages of blastocyst formation, during one of roughly sixteen developmental phases. Therefore, the detection of CS is particularly challenging due to their limited occurrence, thin and low-contrast morphology, variable shapes, and transient presence across focal planes. Furthermore, the substantial class imbalance between CS-negative and CS-positive samples presents an additional challenge for automated analysis, as data-driven models tend to become biased toward the majority class, thereby reducing sensitivity in detecting minority CS instances \cite{ahm00}.

To address the absence of labeled data for CS analysis, we first design a human-in-the-loop annotation pipeline that efficiently streamlines and semi-automates CS labeling in embryo time-lapse images. Building on this curated dataset, we develop a two-stage deep learning framework that (1) determines the presence of CS within each frame and (2) precisely localizes its spatial extent.
To enhance discrimination under such imbalance and feature uncertainty, we propose a novel training objective, the \textbf{N}ovel \textbf{U}ncertainty-aware \textbf{C}ontractive \textbf{E}mbedding (NUCE) loss. NUCE integrates (1) confidence-aware weighting, which emphasizes uncertain and minority samples, with (2) embedding contraction, which encourages compact, well-separated class clusters in the feature space. 
As shown in Fig. \ref{fig:bar}, NUCE consistently improves the F1-score across multiple backbone architectures, highlighting its robustness and generalization capability.
This synergy enhances the robustness and sensitivity to rare CS patterns, ensuring minority instances remain effectively represented during training.
The contributions of this work are as follows:
\begin{enumerate}
    \item We introduce the first computational framework for automated CS detection in human IVF embryo time-lapse images, including a human-in-the-loop annotation pipeline that produces the first expert-validated CS dataset and a two-stage architecture that separates CS presence classification from precise localization.

    \item We propose the NUCE loss, which simultaneously emphasizes uncertain samples and enforces compact, discriminative feature clusters. NUCE delivers consistent performance gains across multiple transformer backbones and, together with RF-DETR–based localization, establishes a state-of-the-art (SOTA) solution for detecting extremely thin, low-contrast CS structures.
\end{enumerate}

\noindent The remainder of this paper is organized as follows: Sec. \ref{sec:relatedwork} reviews the relevant literature and discusses existing approaches related to our task. Sec. \ref{sec:method} details the proposed two-stage framework and the NUCE loss. Sections \ref{sec:exps} and \ref{sec:results} describe the experimental setup and present comprehensive evaluations, including quantitative results and ablation studies. Finally, Sec. \ref{sec:conclusion} concludes the paper by summarizing the key contributions and findings.

\section{Related Work} \label{sec:relatedwork}
\noindent The classification of embryo developmental stages has traditionally relied on manual visual assessment by embryologists, who analyze morphological features at each stage of cell division \cite{Coticchio2025IstanbulConsensus}. This procedure is inherently subjective, requiring considerable expertise and experience, and remains both labor-intensive and time-consuming \cite{Ebner_2023}. Recent advances in deep learning have prompted a paradigm shift toward automated and objective evaluation of embryo development using time-lapse imaging data \cite{alsaad2025deep}. These models aim to assist in assessing embryo quality and predicting implantation potential or live-birth outcomes \cite{POPA2024104054}.

Several studies have demonstrated the efficacy of deep learning approaches in this domain. Nguyen et al. \cite{nguyen2023embryosformer} introduced EmbryosFormer, a deformable transformer-based encoder–decoder architecture designed for embryo stage classification. In this framework, the encoder captures robust visual representations from time-lapse embryo images, while the decoder enforces temporal consistency across developmental sequences. 
Similarly, Leahy et al. \cite{leahy2020} developed a comprehensive deep learning pipeline comprising five CNNs to automate the quantification of key morphokinetic features, including zona pellucida segmentation, fragmentation grading, developmental stage classification, and pronuclei segmentation. Kalyani et al. \cite{kalyani2024deep} employed a hybrid architecture combining ResNet50 with a Gated Recurrent Unit (GRU) to predict blastocyst formation from sequential embryo images. 

Building on these advancements, Kim et al. \cite{kim2024multimodal} proposed a multimodal transformer framework that integrates time-lapse imaging with Electronic Health Records to enhance embryo viability prediction. Similarly, Wang et al. \cite{wang2024generalized} introduced IVFormer, a self-supervised visual–temporal contrastive learning transformer designed to improve embryo representation learning. Their approach jointly encodes static and temporal information through shared visual and temporal encoders to facilitate euploidy ranking and live-birth prediction. Furthermore, Liu et al. \cite{liu2024wise} adopted a self-supervised masked autoencoder strategy to identify whether pairs of images originated from the same embryo across different imaging systems, thereby addressing cross-device generalization challenges.
More recently, Liu et al.\ introduced the Dual-branch Local Feature Fusion Enhanced Transformer (DLT-Embryo), a model specifically designed to improve automated embryo assessment by jointly leveraging local and global information \cite{dlt_embryo}. The architecture incorporates a local-branch module equipped with local self-attention, depthwise convolution, and inter-branch feature fusion to enhance fine-grained feature extraction. In parallel, a global-branch module employs multi-head self-attention to capture long-range contextual dependencies. By integrating these complementary representations across multiple developmental stages, DLT-Embryo aims to achieve more robust and biologically informed embryo classification performance.

Despite these advancements, most existing works for embryo assessment continue to focus predominantly on classification approaches, including morphological grading, morphokinetic profiling, and multimodal integration of image with clinical data. These approaches are based on public or private data annotated for morphokinetic features of the developing embryo. However, none of these approaches explicitly incorporates CS as a discriminative feature, despite growing evidence linking CS to enhanced blastocyst viability, accelerated developmental progression, and increased live-birth rates.
Despite these advancements, most existing works for embryo assessment continue to focus predominantly on classification approaches, including morphological grading \cite{Valera2023}, morphokinetic profiling \cite{Shobha2023}, and multimodal integration of image with clinical data \cite{Wang2024}. These approaches are based on data annotated for morphokinetic or morphological features of the developing embryo. However, none of these approaches explicitly incorporates CS as a discriminative feature, despite growing evidence linking CS to enhanced blastocyst viability, accelerated developmental progression, and increased live-birth rates \cite{[40],[41]}.

In spite of notable progress in automated embryo analysis, no prior work has investigated CS detection, and consequently, no loss-function strategies tailored to this task have been proposed. Automated CS detection must simultaneously address severe class imbalance and the need to discriminate extremely subtle visual features. The pronounced scarcity of CS-positive samples predisposes models to bias toward the majority class, reducing sensitivity and compromising overall detection reliability. In the broader medical imaging domain, however, several categories of loss functions have been proposed to mitigate similar imbalance challenges.
(1) Re-weighting losses, such as class-balanced or balanced cross-entropy losses, adjust the contribution of each class based on frequency to counteract majority-class dominance \cite{ahm_ref2}.
(2) Hard-example–focused losses, including focal-type formulations, place greater emphasis on difficult or minority samples to improve minority-class sensitivity.
(3) Feature-distribution–aware losses, such as Center-Focused Affinity Loss (CFAL), encourage compact intra-class structure and improved separation in the embedding space, thereby enhancing discrimination of subtle minority-class features \cite{ahm_ref01}.
Building on these insights, our proposed loss function is designed not only to alleviate class imbalance but also to increase discriminative power, an essential requirement for reliably distinguishing CS from visually similar morphological patterns.

\begin{figure*}[t]
\centering
\includegraphics[width=0.99\textwidth]{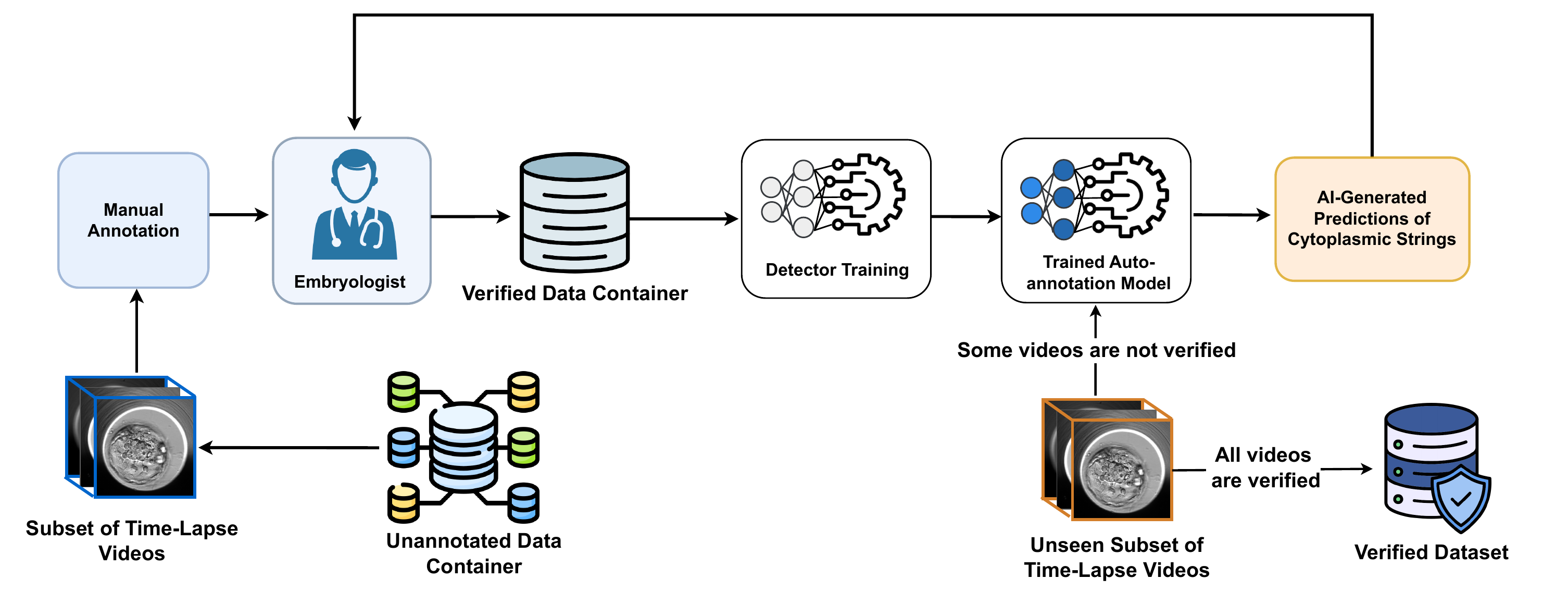}
\caption{
Overview of the annotation pipeline. A subset of time-lapse embryo videos is first manually annotated by expert embryologists, producing a verified data container used to train an automated detector. The trained auto-annotation model then generates cytoplasmic-string predictions for an unseen subset of time-lapse videos. Predicted annotations undergo verification, and all validated outputs are consolidated into a final verified dataset.
}
\label{fig:annotation_pipeline}
\vspace{-4mm}
\end{figure*}

\begin{figure*}[h]
\centering
\includegraphics[width=0.97\linewidth]{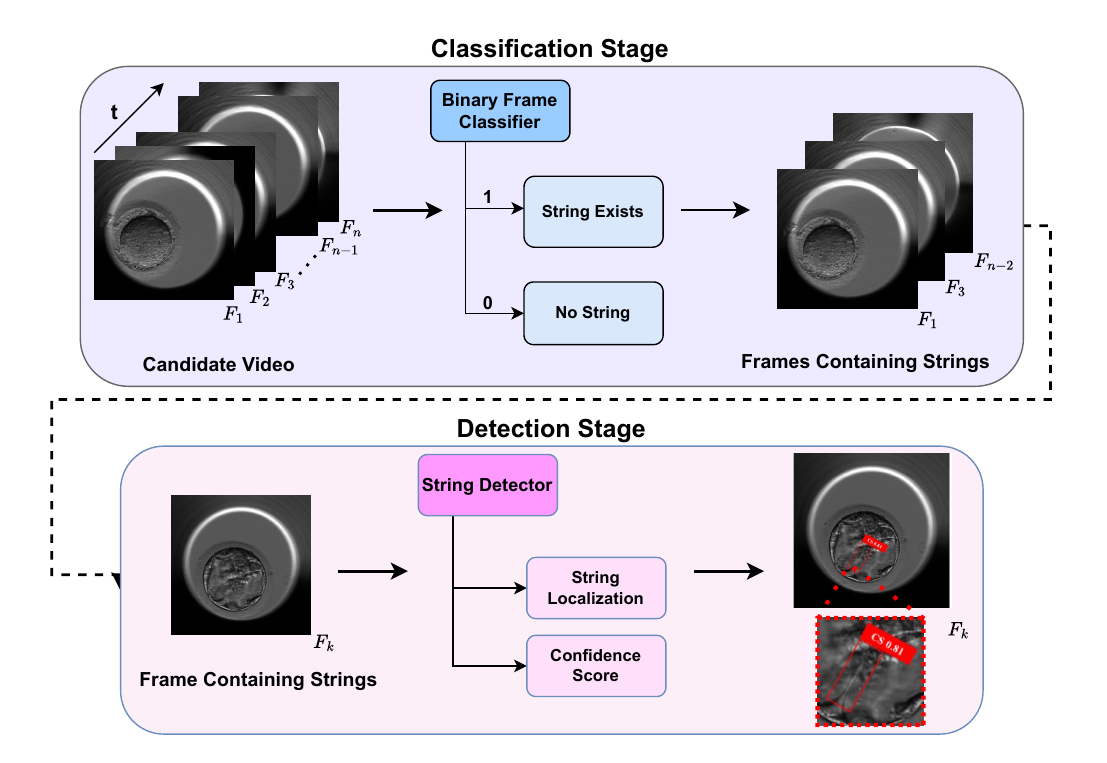}
\caption{Overview of the proposed two-stage framework for CS detection in time-lapse human embryo videos. Stage 1: The classification network identifies CS presence through self-distillation. Stage 2: Localization network detects and localizes CS regions.}
\label{fig:pipeline}
\end{figure*}

\section{Method} \label{sec:method}

\subsection{Overview}
\noindent The proposed framework aims to automate the detection and localization of CS in human embryo time-lapse images. As illustrated in Fig. \ref{fig:pipeline}, the system adopts a two-stage design tailored to address the scarcity and visual subtlety of CS structures. In the first stage, an embryo-level classifier determines the presence of CS within each frame sequence, effectively filtering out CS-negative samples. The second stage then performs fine-grained localization to identify the spatial extent of CS regions within the positive cases.

To enable reliable training, we construct a curated dataset using a human-in-the-loop annotation pipeline that streamlines and semi-automates the labeling of CS structures, as illustrated in Fig. \ref{fig:annotation_pipeline}. The first stage is optimized using the proposed NUCE loss, which mitigates the effects of severe class imbalance and uncertain feature boundaries during CS classification. The second stage is subsequently trained with standard detection objectives, as it only receives CS-positive samples predicted by the first stage. The following subsections describe each component of the framework in detail.

\subsection{Human-in-the-Loop Annotation Pipeline}
\noindent Accurate detection of CS requires high-quality labeled data; however, manual annotation is tedious and error-prone due to the subtle, transient, and low-contrast nature of CS structures. To overcome these limitations, we design a human-in-the-loop annotation pipeline that integrates expert embryologist feedback with automated model suggestions, as illustrated in Fig. \ref{fig:annotation_pipeline}. This strategy accelerates dataset creation while maintaining expert-level accuracy.

\noindent \textbf{Initial model-assisted labeling.}
Raw time-lapse embryo videos, obtained from the publicly available human embryo imaging dataset \cite{[11]}, are first preprocessed to extract representative frames from the blastocyst stage, where CS are most likely to appear. Because no prior dataset existed for this task, a small subset of frames was manually annotated by expert embryologists to serve as a biologically verified support set. A preliminary detection model \cite{rfdetr}, trained on this subset, then produced candidate CS regions represented as probability maps or bounding boxes. These preliminary predictions guided the subsequent expert review process.

\noindent \textbf{Expert review and refinement.}
Embryologists examined each candidate region using an interactive annotation interface that allowed them to accept, modify, or reject the model’s suggestions, and to manually draw additional CS annotations where necessary. Each annotated frame was verified by at least two experts to ensure consistency and minimize inter-observer variability. This combination of model guidance and expert correction effectively balanced annotation speed and biological accuracy.

\noindent \textbf{Iterative refinement.}
Following expert verification, corrected labels were reintegrated into the training set, and the detection model was retrained to generate improved predictions for the remaining unlabeled data. This cycle of prediction, validation, and retraining was repeated until the end of the annotation cycle, substantially reducing manual workload while enhancing labeling precision. Through this iterative process, a general morphokinetic dataset was transformed into a biologically validated CS detection benchmark.

\noindent \textbf{Dataset summary.}
The final dataset comprises 90 developmental sequences with a total of 13,568 frames, among which only 271 contain visible CS instances, highlighting the extreme class imbalance inherent to this task. Each frame includes binary presence labels, and CS-positive samples are further annotated with bounding boxes delineating CS regions. 
\textit{To the best of our knowledge, this represents the first publicly available dataset dedicated to cytoplasmic string detection and localization, establishing a standardized benchmark for future research}.


\subsection{Two-Stage Detection Framework}
\noindent The goal of this study is to automatically detect CS in human embryo time-lapse images under highly imbalanced conditions. To achieve this, we propose a two-stage detection framework, as illustrated in Fig. \ref{fig:pipeline}. Let the dataset be represented as \(\mathcal{D} = \{(x_i, y_i)\}_{i=1}^{N}\) where each image \(x_i \in \mathbb{R}^{H \times W \times C}\) corresponds to a blastocyst-stage frame, and \(y_i\) denotes the associated annotations. If CS are present, \(y_i\) contains one or more bounding boxes \(b_j = (x_{\min}, y_{\min}, x_{\max}, y_{\max})\) with the binary class label \(c_j \in \{\text{CS}, \text{Background}\}\); otherwise, \(y_i = \varnothing\).

Given the extreme class imbalance, where only a small fraction of frames contain visible CS, we adopt a two-stage architecture that separates classification and localization. The first stage performs embryo-level classification, filtering out CS-negative samples to reduce false positives and computational overhead. The second stage performs region-level localization within the remaining CS-positive frames to identify the precise spatial extent of CS structures.

Formally, let \(g_{\phi}(\cdot)\) denote the classification network parameterized by \(\phi\). For an input frame \(x_i\), the predicted probability of CS presence is \(\hat{p}_i = g_{\phi}(x_i)\), where \(\hat{p}_i \in [0,1]\) represents the CS-positive confidence. Frames with \(\hat{p}_i \geq \tau\) are treated as CS-positive and passed to the detection stage:
\begin{equation}
\hat{y}_i =
\begin{cases}
1, & \text{if } \hat{p}_i \geq \tau, \\
0, & \text{otherwise.}
\end{cases}
\end{equation}
The first stage is optimized using the proposed NUCE loss to improve discrimination under imbalance and feature uncertainty. 
The second stage is trained with a standard detection objective to localize CS regions within CS-positive samples identified by the classifier. 
This hierarchical design enables the detector to focus solely on informative, CS-relevant frames, thereby improving both computational efficiency and detection sensitivity.

The proposed NUCE loss draws on two established paradigms: (i) uncertainty-aware weighting of ambiguous samples, as seen in focal-type losses \cite{focal,NIPS2017_2650d608}, and (ii) prototype-based embedding regularization, such as Center Loss or Prototype Loss \cite{center_loss}. NUCE integrates these concepts by up-weighting uncertain CS-positive samples—critical under severe imbalance—while contractively aligning embeddings with class anchors to maintain discriminative latent structure. This provides a principled mechanism for dealing with uncertain samples and improving reliability in CS detection.

\subsection{Novel Uncertainty-aware Contractive Embedding}
\noindent We introduce NUCE loss, a training objective designed to enhance the robustness and discriminative power of deep classification models. 
This loss is applied to the classification stage of the proposed framework to address severe class imbalance and feature uncertainty.
NUCE simultaneously addresses two critical challenges in representation learning: the need to focus learning on uncertain or ambiguous examples, and the requirement to maintain compact, well-separated class clusters in the latent space. This is achieved through a principled combination of confidence-aware sample reweighting and a geometric contraction term that aligns features with class-specific anchors.

Let a mini-batch of $B$ training samples be denoted as \( \mathcal{B} = \{(\mathbf{x}_i, y_i)\}_{i=1}^B \), where \( \mathbf{x}_i \in \mathbb{R}^n \) is an input instance and \( y_i \in \{1, \dots, K\} \) is its associated class label from among \(K\) possible classes. Each input is passed through a feature extractor network \( f_\theta \) to produce a feature embedding \( \mathbf{h}_i = f_\theta(\mathbf{x}_i) \in \mathbb{R}^d \). We collect these into a matrix \( H \in \mathbb{R}^{B \times d} \), where each row corresponds to a sample's feature vector. 
Explicitly, the feature matrix is defined as:
\begin{equation}
H = 
\begin{bmatrix}
h_{1,1} & h_{1,2} & \cdots & h_{1,d} \\
h_{2,1} & h_{2,2} & \cdots & h_{2,d} \\
\vdots & \vdots & \ddots & \vdots \\
h_{B,1} & h_{B,2} & \cdots & h_{B,d}
\end{bmatrix},
\end{equation}
where \( h_{i,j} \) denotes the $j^{th}$ component of the feature vector for the $i^{th}$ sample.
To compute class predictions, each feature vector is projected via a linear classifier with weight matrix \( W \in \mathbb{R}^{K \times d} \), producing a logit vector \( \mathbf{u}_i = W \mathbf{h}_i \in \mathbb{R}^K \). 
When computed over the entire batch, the logits are assembled into a matrix:
\begin{equation}
U = H W^\top,
\end{equation}
where \( U \in \mathbb{R}^{B \times K} \). 
These logits are then passed through a softmax activation to obtain class probability distributions for each sample:
\begin{equation}
P = \operatorname{softmax}(U),
\end{equation}
where each row \( \mathbf{p}_i \in \mathbb{R}^K \) corresponds to the predicted probabilities for sample \( i \).
Ground truth labels are represented using a one-hot encoding matrix \( Y \in \{0,1\}^{B \times K} \), where the $i^{th}$ row has a 1 in the position corresponding to $y_i$ and 0 elsewhere. Furthermore, to model class structure within the embedding space, we define a learnable anchor matrix \( A \in \mathbb{R}^{K \times d} \), where each row \( \mathbf{a}_k \) represents the prototype or centroid of class \( k \). 
This anchor matrix is defined as:
\begin{equation}
A = 
\begin{bmatrix}
a_{1,1} & a_{1,2} & \cdots & a_{1,d} \\
a_{2,1} & a_{2,2} & \cdots & a_{2,d} \\
\vdots & \vdots & \ddots & \vdots \\
a_{K,1} & a_{K,2} & \cdots & a_{K,d}
\end{bmatrix}.
\end{equation}
To modulate each sample's influence based on its predicted confidence, we define an uncertainty-aware weight for each sample. The idea is to increase the loss contribution of ambiguous examples (e.g., those near decision boundaries), while down-weighting confidently classified ones. 
This is formulated as:
\begin{equation}
\omega_i = \left( 1 - \max_k p_{i,k} \right)^\gamma,
\end{equation}
where \( \gamma \ge 0 \) is a tunable parameter that adjusts the emphasis on uncertain samples. 
The classification loss is then defined using these weights to scale the standard log-likelihood loss:
\begin{equation}
\mathcal{L}_{\text{risk}} = -\frac{1}{B} \sum_{i=1}^B \omega_i \log p_{i, y_i},
\end{equation}
where \( p_{i, y_i} \) is the predicted probability for the correct class of sample $i$.
In parallel, we apply a contraction-based regularization to ensure that samples of the same class are pulled closer to their corresponding anchor in the latent space. This regularization helps form tighter, well-separated class clusters and improves generalization under noisy or imbalanced conditions. The contractive term is defined as:
\begin{equation}
\mathcal{L}_{\text{contract}} = \frac{1}{2B} \sum_{i=1}^B \left\| \mathbf{h}_i - \mathbf{a}_{y_i} \right\|_2^2,
\end{equation}
which computes the squared distance between each feature and its associated anchor and averages it over the batch.
We combine both terms into the final NUCE loss function:
\begin{equation}
\mathcal{L}_{\text{NUCE}} = \lambda_r \cdot \mathcal{L}_{\text{risk}} + \lambda_c \cdot \mathcal{L}_{\text{contract}},
\end{equation}
where \( \lambda_r \) and \( \lambda_c \) are positive scalar hyperparameters that balance the importance of each component.
To facilitate efficient vectorized computation, we reformulate the same objective in matrix form. Let \( \bm{\omega} \in \mathbb{R}^B \) be the vector of per-sample weights. The classification loss becomes:
\begin{equation}
\mathcal{L}_{\text{risk}} = -\frac{1}{B} \, \bm{\omega}^\top \, \operatorname{diag}(Y P^\top),
\end{equation}
where \( \operatorname{diag}(Y P^\top) \in \mathbb{R}^B \) extracts the predicted probabilities for the correct classes. Likewise, the contractive loss is expressed using the Frobenius norm:
\begin{equation}
\mathcal{L}_{\text{contract}} = \frac{1}{2B} \left\| H - Y A \right\|_F^2.
\end{equation}
Consequently, the full NUCE objective in matrix form is:
\begin{equation}
\mathcal{L}_{\text{NUCE}} = \lambda_r \left( -\frac{1}{B} \bm{\omega}^\top \operatorname{diag}(Y P^\top) \right) + \lambda_c \left( \frac{1}{2B} \left\| H - Y A \right\|_F^2 \right),
\end{equation}
offering a cohesive formulation that improves both predictive reliability and latent structure without requiring additional architectural modifications.

\section{Experimental Setup} \label{sec:exps}

\subsection{Implementation Details}
\noindent All experiments were implemented in the PyTorch framework and executed on a workstation equipped with a single NVIDIA RTX 3080 GPU (8 GB VRAM). 
Both the classification and detection models were fine-tuned from publicly available pretrained weights. 
Each model was trained for 10 epochs using the Adam optimizer with an initial learning rate of \(10^{-3}\) and a batch size of 128. 
A cosine learning rate schedule and early stopping based on validation performance were employed to prevent overfitting.

For the classification stage, the NUCE loss was applied with weighting parameters \(\lambda_r = \text{1.0}\), \(\lambda_c = \text{0.5}\), and uncertainty exponent \(\gamma = \text{2}\). 
The detection stage was trained using standard classification and bounding-box regression losses with foreground–background balancing to account for the limited number of CS-positive samples. 
All results were averaged over three independent runs with different random seeds. 
For evaluation, we adopted a \(K\)-fold cross-validation protocol at the embryo level to prevent data leakage between developmental sequences. 
In each fold, embryos were partitioned into mutually exclusive training and validation subsets, and we report the mean performance across all folds.
Unless otherwise stated, all reported metrics correspond to the mean performance across \(K\)-fold cross-validation.

\begin{table*}[t!]
\centering
\caption{Comparison of classification models with different loss functions. 
For each metric, the left column shows the macro average, and the right column shows the weighted average.}
\vspace{-0.7em}
\label{tab:model_loss_comparison}
\resizebox{0.99\linewidth}{!}{
\begin{tabular}{@{}llcccccccc@{}}
\specialrule{2pt}{0pt}{0pt}
\multirow{3}{*}{\textbf{Model}} & 
\multirow{3}{*}{\textbf{Loss}} & 
\multirow{3}{*}{\textbf{Accuracy}} & 
\multicolumn{2}{c}{\textbf{F1-score}} & 
\multicolumn{2}{c}{\textbf{Precision}} & 
\multicolumn{2}{c}{\textbf{Recall}} \\ 
\cmidrule(lr){4-5} \cmidrule(lr){6-7} \cmidrule(lr){8-9}
& & & Macro & Weighted & Macro & Weighted & Macro & Weighted \\ 
\specialrule{2pt}{0pt}{0pt}

\multirow{5}{*}{\textbf{ViT-B} \cite{vit}} & Cross Entropy & 82.9 & 78.9 & 82.6 & 77.4 & 82.9 & 78.1 & 82.7 \\ 
& Focal Loss & 80.2 & 78.4 & 79.5 & 68.3 & 80.2 & 70.6 & 78.1 \\
& Center Loss & 84.8 & 80.8 & 85.1 & 81.8 & 84.8 & 81.3 & 84.9 \\
& Affinity Loss & 72.4 & 36.2 & 52.4 & 50.0 & 72.4 & 41.9 & 60.7 \\
& NUCE Loss & \textbf{88.5} \tiny{(\textcolor{red}{\textbf{3.7$\uparrow$}})} & \textbf{85.1} \tiny{(\textcolor{red}{\textbf{4.3$\uparrow$}})} & \textbf{89.1} \tiny{(\textcolor{red}{\textbf{4.0$\uparrow$}})} & \textbf{87.4} \tiny{(\textcolor{red}{\textbf{5.6$\uparrow$}})} & \textbf{88.5} \tiny{(\textcolor{red}{\textbf{3.7$\uparrow$}})} & \textbf{86.1} \tiny{(\textcolor{red}{\textbf{4.8$\uparrow$}})} & \textbf{88.7} \tiny{(\textcolor{red}{\textbf{3.8$\uparrow$}})}
\\

\hline

\multirow{5}{*}{\textbf{Swin-B} \cite{swin}} & Cross Entropy & 82.0 & 78.8 & 81.3 & 73.7 & 82.0 & 75.5 & 81.2 \\ 
& Focal Loss & 77.4 & 84.4 & 80.9 & 59.7 & 77.4 & 59.7 & 71.6 \\
& Center Loss & 83.9 & 81.6 & 83.3 & 75.9 & 83.9 & 78.0 & 83.1 \\
& Affinity Loss & 84.2 & 82.1 & 85.5 & 84.2 & 86.3 & 83.7 & 86.8 \\
& NUCE Loss & \textbf{88.9} \tiny{(\textcolor{red}{\textbf{4.7$\uparrow$}})} & \textbf{86.0} \tiny{(\textcolor{red}{\textbf{1.6$\uparrow$}})} & \textbf{89.1} \tiny{(\textcolor{red}{\textbf{3.6$\uparrow$}})} & \textbf{86.7} \tiny{(\textcolor{red}{\textbf{2.5$\uparrow$}})} & \textbf{88.9} \tiny{(\textcolor{red}{\textbf{2.6$\uparrow$}})} & \textbf{86.3} \tiny{(\textcolor{red}{\textbf{2.6$\uparrow$}})} & \textbf{89.0} \tiny{(\textcolor{red}{\textbf{2.2$\uparrow$}})}
\\

\hline

\multirow{5}{*}{\textbf{DeiT-B} \cite{deit}} & Cross Entropy & 85.3 & 82.7 & 84.8 & 79.0 & 85.3 & 80.5 & 84.8 \\ 
& Focal Loss & 84.3 & 85.9 & 84.9 & 73.7 & 84.3 & 76.9 & 82.8 \\
& Center Loss & 86.6 & 83.6 & 86.5 & 82.5 & 86.6 & 83.0 & 86.5 \\
& Affinity Loss & 72.4 & 36.2 & 52.4 & 50.0 & 72.4 & 41.9 & 60.7 \\
& NUCE Loss & \textbf{89.9} \tiny{(\textcolor{red}{\textbf{3.3$\uparrow$}})} & \textbf{87.6} \tiny{(\textcolor{red}{\textbf{1.7$\uparrow$}})} & \textbf{89.8} \tiny{(\textcolor{red}{\textbf{3.3$\uparrow$}})} & \textbf{86.8} \tiny{(\textcolor{red}{\textbf{4.3$\uparrow$}})} & \textbf{89.9} \tiny{(\textcolor{red}{\textbf{3.3$\uparrow$}})} & \textbf{87.2} \tiny{(\textcolor{red}{\textbf{4.2$\uparrow$}})} & \textbf{89.8} \tiny{(\textcolor{red}{\textbf{3.3$\uparrow$}})} 
\\

\hline

\multirow{5}{*}{\textbf{DINOv2-B} \cite{dinov2}} & Cross Entropy & 86.2 & 83.4 & 85.3 & 80.1 & 85.8 & 81.6 & 85.7 \\ 
& Focal Loss & 86.2 & 85.4 & 85.9 & 78.6 & 86.2 & 81.0 & 85.5 \\
& Center Loss & 86.2 & 86.0 & 86.1 & 78.1 & 86.2 & 80.8 & 85.3 \\
& Affinity Loss & 72.4 & 36.2 & 52.4 & 50.0 & 72.4 & 41.9 & 60.7 \\
& NUCE Loss & \textbf{91.2} \tiny{(\textcolor{red}{\textbf{5.0$\uparrow$}})} & \textbf{89.2} \tiny{(\textcolor{red}{\textbf{3.2$\uparrow$}})} & \textbf{91.2} \tiny{(\textcolor{red}{\textbf{5.1$\uparrow$}})} & \textbf{88.8} \tiny{(\textcolor{red}{\textbf{8.7$\uparrow$}})} & \textbf{91.2} \tiny{(\textcolor{red}{\textbf{5.0$\uparrow$}})} & \textbf{89.0} \tiny{(\textcolor{red}{\textbf{7.4$\uparrow$}})} & \textbf{91.2} \tiny{(\textcolor{red}{\textbf{5.5$\uparrow$}})}
\\

\hline

\multirow{5}{*}{\textbf{CLIP-B/32} \cite{clip}} & Cross Entropy & 76.5 & 78.1 & 77.3 & 59.0 & 76.5 & 58.9 & 70.9 \\ 
& Focal Loss & 74.7 & 80.0 & 77.5 & 54.7 & 74.7 & 51.5 & 66.5 \\
& Center Loss & 81.1 & 81.8 & 81.4 & 68.4 & 81.1 & 71.0 & 78.7 \\
& Affinity Loss & 86.2 & 83.7 & 85.8 & 80.7 & 86.2 & 81.9 & 85.8 \\
& NUCE Loss & \textbf{90.8} \tiny{(\textcolor{red}{\textbf{4.6$\uparrow$}})} & \textbf{88.0} \tiny{(\textcolor{red}{\textbf{4.3$\uparrow$}})} & \textbf{91.0} \tiny{(\textcolor{red}{\textbf{5.2$\uparrow$}})} & \textbf{89.5} \tiny{(\textcolor{red}{\textbf{8.8$\uparrow$}})} & \textbf{90.8} \tiny{(\textcolor{red}{\textbf{4.6$\uparrow$}})} & \textbf{88.7} \tiny{(\textcolor{red}{\textbf{6.8$\uparrow$}})} & \textbf{90.9} \tiny{(\textcolor{red}{\textbf{5.1$\uparrow$}})}
\\

\specialrule{2pt}{0pt}{0pt}
\end{tabular}
}
\end{table*}

\subsection{Adopted Methods}
\noindent \textbf{Stage One: Classification Methods.}
We evaluated five representative Vision Transformer (ViT) variants as backbone classifiers for CS presence prediction: ViT-B \cite{vit}, Swin-B \cite{swin}, DeiT-B \cite{deit}, DINOv2-B \cite{dinov2}, and CLIP-B/32 \cite{clip}. 
These architectures collectively cover standard, hierarchical, data-efficient, self-supervised, and vision–language pretrained transformer designs. 

\noindent \textbf{Loss Functions.}
To establish comprehensive baselines for evaluating the proposed NUCE loss, each classifier was trained with several widely adopted objectives: 
Cross-Entropy loss, Focal loss \cite{focal}, Center loss \cite{center_loss}, and Affinity loss \cite{affinity}. 
These losses represent standard, imbalance-aware, and embedding-regularized paradigms, enabling a broad comparison of classification performance under different optimization strategies.

\noindent \textbf{Stage Two: Detection Methods.}
For the localization stage, we benchmarked six SOTA object detection architectures encompassing both transformer-based and convolution-based paradigms: 
RF-DETR \cite{rfdetr}, RT-DETR \cite{rtdetr}, RT-DETRv2 \cite{rtdetrv2}, YOLOv8 \cite{yolov8}, YOLOv11 \cite{yolov11}, and YOLOv12 \cite{yolov12}. 
This selection spans recent DETR-style transformer detectors emphasizing query refinement and real-time performance, as well as modern YOLO variants designed for high-accuracy detection of small and low-contrast objects.

\subsection{Evaluation Metrics}
\noindent To ensure a fair and comprehensive assessment of both stages in the proposed framework, we employ standard evaluation metrics separately for the classification and detection tasks.

\noindent \textbf{Stage One: Classification.}
The performance of the classification stage is quantified using four widely adopted metrics: accuracy, precision, recall, and F1-score. 
Let TP, TN, FP, and FN denote the numbers of true positives, true negatives, false positives, and false negatives, respectively. 
Accuracy, precision, and recall are defined as:
\begin{equation}
\begin{aligned}
\text{Accuracy} &= \frac{\text{TP} + \text{TN}}{\text{TP} + \text{TN} + \text{FP} + \text{FN}}, \\
\text{Precision} &= \frac{\text{TP}}{\text{TP} + \text{FP}}, \quad
\text{Recall} = \frac{\text{TP}}{\text{TP} + \text{FN}}.
\end{aligned}
\end{equation}
The F1-score, which represents the harmonic mean of precision and recall, is computed as:
\begin{equation}
\text{F1-score} = 2 \times \frac{\text{Precision} \times \text{Recall}}{\text{Precision} + \text{Recall}}.
\end{equation}
\noindent To account for the severe class imbalance present in the dataset, we report both macro and weighted variants of all metrics.
Macro averaging assigns equal weight to each class, providing insight into performance across minority classes, while weighted averaging incorporates class frequencies, reflecting overall performance more faithfully.
The F1-score serves as the primary evaluation criterion for the classification stage, as it offers a balanced measure of model performance under imbalance.

\noindent \textbf{Stage Two: Detection.}
For the detection stage, model performance is evaluated using mean Average Precision (mAP), the standard metric for object detection tasks. 
The Average Precision (AP) is computed as the area under the precision–recall curve for each class, and mAP represents the mean of these values across all classes. 
Following common practice, we report mAP at Intersection-over-Union (IoU) thresholds of 0.25 (\text{mAP@0.25}), 0.5 (\text{mAP@0.5}), and 0.75 (\text{mAP@0.75}). 
Because the CS detection task involves a single object class, the reported mAP values directly reflect the precision–recall performance of CS localization. 
All metrics are computed at the frame level and averaged over three independent runs to ensure statistical robustness.

\begin{figure*}[t]
    \centering
    \includegraphics[width=0.99\textwidth]{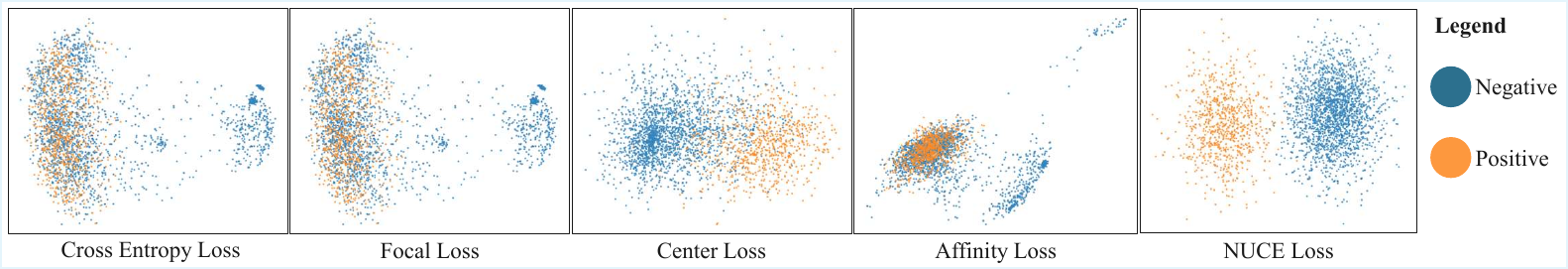}
    \caption{Comparison of learned embedding distributions across loss functions. 
    The high-dimensional feature embeddings extracted from the ViT-B final layer are projected into a 2D space using Principal Component Analysis (PCA) to visualize the structural differences induced by each loss function.}
    \label{fig:loss_comparison}
\end{figure*}

\begin{figure*}[t]
    \centering
    \includegraphics[width=0.8\textwidth]{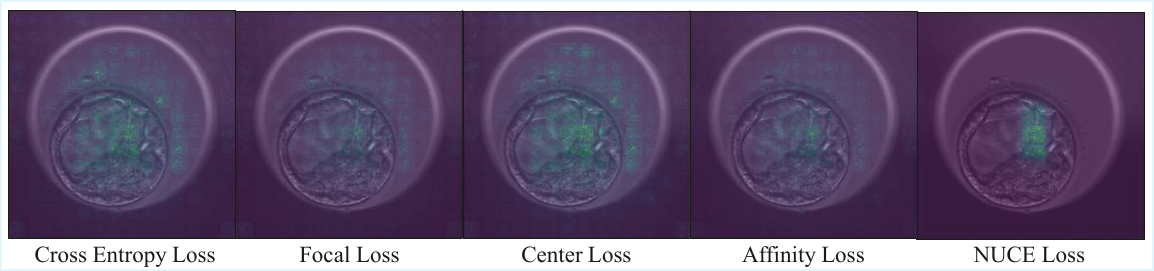}
    \caption{Class-discriminative heatmaps generated under different loss functions.}
    \label{fig:heatmaps}
\end{figure*}

\section{Results} \label{sec:results}

\begin{figure*}[h]
\centering
\includegraphics[width=0.99\linewidth]{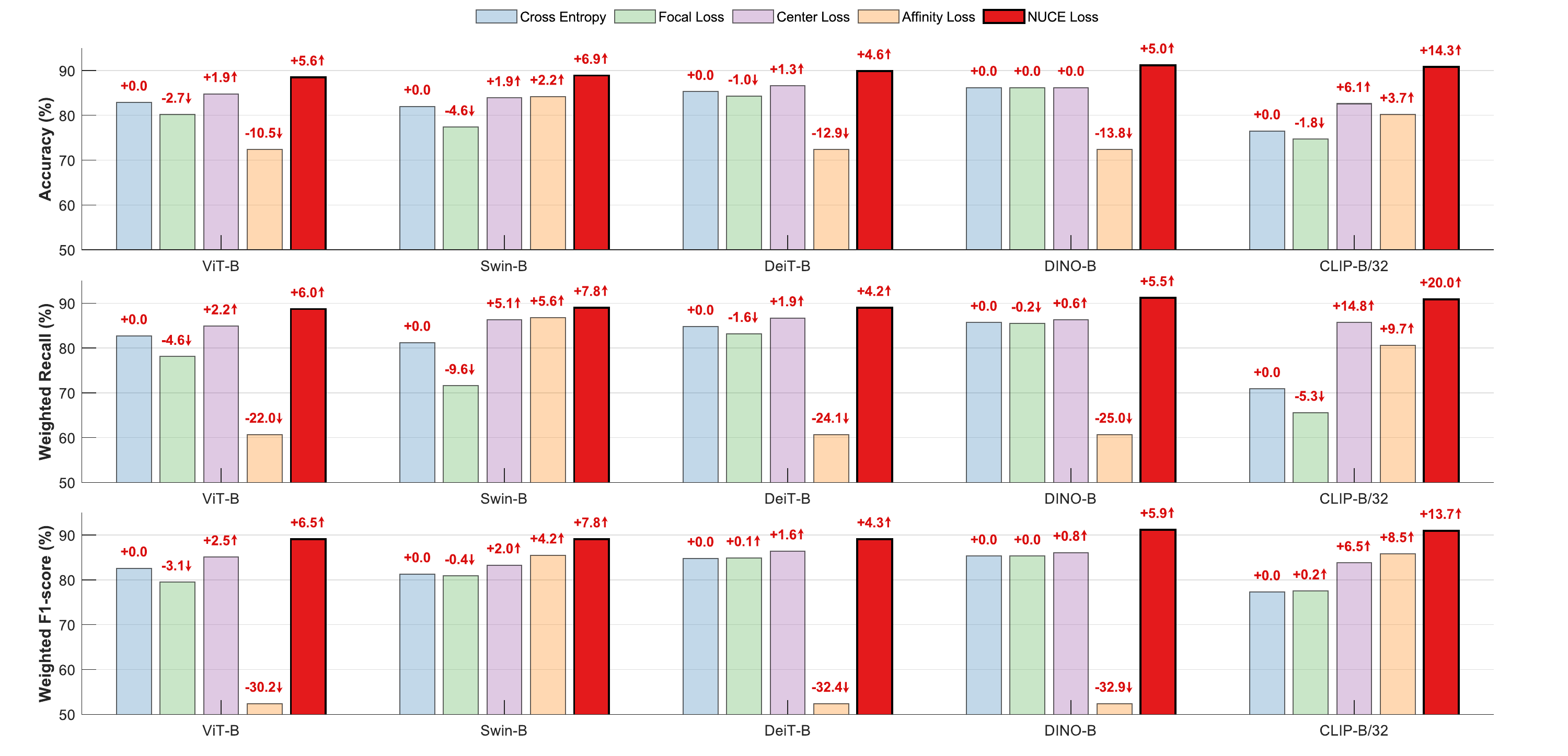}
\caption{Quantitative comparison of classification models trained with different loss functions considering the Cross Entropy as reference. The results show consistent improvements in accuracy, weighted recall, and weighted F1-score with the proposed NUCE loss.}
\label{fig:bar1}
\end{figure*}

\subsection{Stage 1: Classification}
\noindent The first stage assesses the ability of transformer-based classifiers to distinguish CS-positive from CS-negative frames under severe class imbalance. Five backbone architectures were trained with five loss functions, and their performance is reported in Table \ref{tab:model_loss_comparison}. The goal is to determine whether the proposed NUCE objective improves sensitivity and discriminative consistency compared to standard optimization approaches.
Across all models, NUCE consistently achieves the highest accuracy, precision, recall, and F1-score. Compared to Cross-Entropy, Focal, Center, and Affinity losses, NUCE exhibits a clear performance margin, a trend also visible in the aggregated metrics of Fig. \ref{fig:bar1} and the performance plot in perfor.svg. These plots show a uniform upward shift in accuracy, weighted recall, and weighted F1, confirming that NUCE improves both minority-class recognition and overall decision reliability.

\noindent \textbf{Backbone-level performance analysis.}
For ViT-B and Swin-B, NUCE raises accuracy by 3.7–4.7 points, while boosting macro-F1 by 4.3 and 1.6 points, respectively. DeiT-B and CLIP-B/32 also show consistent gains, with NUCE improving accuracy by 3.3–4.6 points. The strongest effect is observed in DINOv2-B, where accuracy increases by 5.0 points, and weighted precision jumps by 8.7 points, confirming that NUCE synergizes especially well with self-supervised transformer representations.

\noindent \textbf{Embedding-space interpretation.}
The embedding visualizations from Fig. \ref{fig:loss_comparison} reveal the structural differences in the learned feature space using ViT-B as a backbone. Under Cross-Entropy, Focal, and Center losses, positive and negative samples remain partially interwoven, and Affinity loss collapses minority-class representation altogether. NUCE produces compact, well-separated clusters, validating the impact of its contractive embedding term and uncertainty-aware reweighting. 

\noindent \textbf{Heatmap-based qualitative analysis.}
To further examine how each loss function influences spatial attention, we generated class-discriminative heatmaps using ViT-B as a backbone, shown in Fig. \ref{fig:heatmaps}.
Across baseline losses, the highlighted regions are diffuse, unstable, or misaligned with the actual CS-bearing areas—especially under Focal and Affinity losses, where the model fixates on irrelevant background structures. Cross-Entropy and Center Loss offer slightly more coherent activations but remain scattered.
NUCE, in contrast, produces sharp and consistent activation along the blastocoel cavity region, where CS structures naturally form. The heatmap demonstrates that the model is learning the correct morphological cues rather than relying on background intensity artifacts. This qualitative behavior is aligned with the numerical gains and embedding-space separation, confirming that NUCE enhances both where and how the model looks when predicting CS presence.

\noindent \textbf{Discussion.}
These results show that NUCE significantly enhances classifier robustness, particularly in the presence of subtle morphological cues and extreme class imbalance. NUCE not only increases numerical performance but also yields a more structured, discriminative latent space, crucial for reliably identifying rare CS-positive frames that feed into the detection stage.

\begin{table}[h!]
\centering
\caption{Detection performance of YOLO, RT-DETR, RT-DETRv2, and RF-DETR architectures on CS localization.}
\label{tab:detection}
\vspace{-0.8em}
        
\begin{tabular}{@{}llcccc@{}}
\specialrule{2pt}{0pt}{0pt}
\textbf{Model} & 
\textbf{Size} & 
\textbf{mAP} & 
\textbf{mAP@25} & 
\textbf{mAP@50} & 
\textbf{mAP@75} 
\\ 

\specialrule{2pt}{0pt}{0pt}

\multirow{5}{*}{\textbf{YOLOv8} \cite{yolov8}}
& n & 14 & 32 & 24 & 16 \\
& s & 4 & 20 & 12 & 1  \\
& m & 6 & 22 & 15 & 3  \\
& l & 7 & 24 & 18 & 4  \\
& x & 8 & 26 & 20 & 5  \\
\hline

\multirow{5}{*}{\textbf{YOLOv11} \cite{yolov11}}
& n & 15 & 34 & 26 & 17 \\
& s & 6 & 22 & 14 & 3  \\
& m & 8 & 25 & 18 & 5  \\
& l & 9 & 27 & 21 & 6  \\
& x & 10 & 29 & 23 & 7  \\
\hline

\hline

\multirow{5}{*}{\textbf{YOLOv12} \cite{yolov12}}
& n & 4 & 18 & 14 & 1 \\
& s & 5 & 20 & 9  & 4 \\
& m & 5 & 22 & 20 & 1 \\
& l & 5 & 21 & 15 & 1 \\
& x & 5 & 23 & 16 & 1 \\
\hline

\multirow{2}{*}{\textbf{RT-DETR} \cite{rtdetr}}
& l & 40 & 88 & 80 & 30 \\
& x & 38 & 86 & 77 & 28 \\
\hline

\multirow{3}{*}{\textbf{RT-DETRv2} \cite{rtdetrv2}}
& s & 28 & 72 & 60 & 20 \\
& m & 33 & 78 & 68 & 24 \\
& l & 38 & 84 & 75 & 28 \\
\hline

\multirow{5}{*}{\textbf{RF-DETR \cite{rfdetr}}}
& n & 30.4 & 82.3 & 70.1 & 25.3 \\
& s & 38.9 & 88.5 & 80.4 & 33.7 \\
& m & \textbf{45.7} & \textbf{93.1} & \textbf{89.5} & \textbf{39.7} \\
& b & 43.1 & 90.4 & 86.9 & 37.5 \\
& l & 44.6 & 91.8 & 88.1 & 38.6 \\

\specialrule{2pt}{0pt}{0pt}
\end{tabular}
\end{table}

\noindent The second stage focuses on localizing CSs within CS-positive frames selected by the classifier. Five representative detection families—YOLOv8 \cite{yolov8}, YOLOv11 \cite{yolov11}, YOLOv12 \cite{yolov12}, RT-DETR \cite{rtdetr}, RT-DETRv2 \cite{rtdetrv2}, and RF-DETR \cite{rfdetr}—were trained and evaluated on the CS-positive subset. 
Performance was measured using mAP, mAP@25, mAP@50, and mAP@75 (Table \ref{tab:detection}), where mAP@25 provides an additional low-IoU perspective suited for extremely thin and low-contrast structures such as CS.

\subsection{Stage 2: Detection}
\noindent The second stage focuses on localizing CSs within CS-positive frames selected by the classifier. Five representative detection families, YOLOv8 \cite{yolov8}, YOLOv11 \cite{yolov11}, YOLOv12 \cite{yolov12}, RT-DETR \cite{rtdetr}, RT-DETRv2 \cite{rtdetrv2}, and RF-DETR \cite{rfdetr}, were trained and evaluated on the CS-positive subset. Performance was measured using mAP, mAP@25, mAP@50, and mAP@75 (Table \ref{tab:detection}). mAP@25 provides an additional low-IoU perspective suited for CS.
The goal is to determine which architecture can handle the extremely subtle, thin, low-contrast CS structures.

RF-DETR provides the strongest detection performance by a large margin. While all YOLO variants struggle to exceed mAP@50 values above 26\% and show limited gains even at the more permissive mAP@25 threshold, the RT-DETR models reach 60–80\% at IoU=0.5 and 70–88\% at IoU=0.25. 
However, RF-DETR substantially surpasses all of them. The RF-DETR-m variant achieves a leading performance of 93.1 mAP@25, 45.7 mAP, 89.5 mAP@50, and 39.7 mAP@75, establishing a clear SOTA for CS localization. 
This is expected, given the architectural emphasis on multi-scale refinement and query-based attention, features that are crucial for extremely thin anatomical structures.

\noindent \textbf{Detector-level performance analysis.}

\noindent \emph{(a) YOLO variants.}
All YOLO models—v8, v11, and v12—perform poorly across all metrics, including mAP@25, with values ranging only from 18 to 34. Although the lower IoU threshold yields modest improvements compared to mAP@50 and mAP@75, the gains remain insufficient for reliable CS localization. Their convolution-centric design and limited global receptive field make them unsuitable for detecting fine, filament-like structures embedded in low-contrast regions.

\noindent \emph{(b) RT-DETR and RT-DETRv2.}
RT-DETR and RT-DETRv2 show moderate success, with mAP@25 values between 72 and 88 and mAP@50 reaching 60–80\% for larger variants. However, their high-IoU performance (mAP@75) remains low (20–30\%), indicating difficulty in precisely bounding extremely thin CS structures. 
These results suggest that although transformer attention improves coarse localization (reflected in strong mAP@25 and mAP@50), a single-pass transformer remains insufficient for high-precision boundary estimation.

\begin{figure*}[t]
    \centering
    \includegraphics[width=0.99\textwidth]{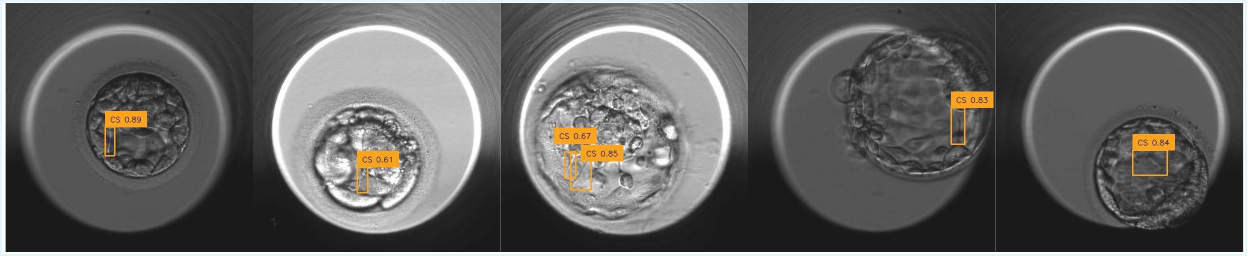}
    \caption{Qualitative CS localization results across test samples.}
    \label{fig:detection_results}
\end{figure*}

\noindent \textbf{RF-DETR.}
RF-DETR consistently dominates all baselines. The medium, large, and base models achieve the highest scores across all IoU thresholds, including the strongest performance at mAP@25. 
Their refinement-based transformer design enables progressive feature aggregation, amplifying the weak spatial signatures of CS. 
The strong performance at high IoU (mAP@75) further highlights their superior ability to produce tight, spatially consistent bounding boxes.
Qualitative examples of RF-DETR predictions are shown in Fig.~\ref{fig:detection_results}, illustrating the model's ability to localize extremely thin CS filaments under low-contrast conditions.

\noindent \textbf{Discussion.}
The large performance gap between RF-DETR and the YOLO families indicates that shallow, convolution-dominated architectures are insufficient for CS localization. The thin, subpixel-like, and low-contrast nature of CS requires multi-scale attention, iterative refinement, and global spatial reasoning—capabilities inherent in transformer-based detectors. 
Moreover, the contrast between RT-DETR and RF-DETR highlights that refinement stages are essential for reliably capturing micro-level structures, since coarse single-pass predictions (even when strong at mAP@25) are inadequate for high-precision localization required at higher IoU thresholds.

\subsection{Ablation Study}
\noindent To quantify the contribution of each component of the proposed NUCE loss, we conducted an ablation study using ViT-B as the backbone classifier. NUCE consists of two functional elements: (1) an uncertainty-aware weighting term that increases the contribution of ambiguous samples, and (2) a contractive embedding regularizer that pulls features toward their corresponding class anchors. Table \ref{tab:ablation} reports the results of progressively adding these components to a Cross-Entropy baseline.

\noindent \textbf{Effect of uncertainty-aware weighting.}
Incorporating uncertainty weights yields consistent improvements across all metrics. Accuracy increases from 82.9\% to 84.2\%, while F1-score improves from 78.9\% to 82.0\%. This demonstrates that prioritizing ambiguous samples helps counteract the severe class imbalance and encourages the model to reduce confusion around CS-positive frames.

\begin{table}[t]
\centering
\caption{Ablation study of NUCE components on ViT-B. 
Each row adds one component to the baseline CE loss. 
The full NUCE formulation (uncertainty weighting + contractive embedding) achieves the best performance.}
\label{tab:ablation}
\vspace{-0.5em}
\begin{tabular}{@{}lcccc@{}}
\specialrule{2pt}{0pt}{0pt}
\textbf{Configuration} &
\textbf{Accuracy} &
\textbf{Precision} &
\textbf{Recall} &
\textbf{F1-score} \\
\specialrule{2pt}{0pt}{0pt}

Cross-Entropy &
82.9 & 77.4 & 78.1 & 78.9 \\

+ Uncertainty Weighting &
84.2 & 79.8 & 81.7 & 82.0 \\

\begin{tabular}[c]{@{}c@{}}\textbf{Full NUCE}\\\textbf{( + Contractive Embedding )}\end{tabular} &
\textbf{88.5} & \textbf{87.4} & \textbf{86.1} & \textbf{85.1} \\

\specialrule{2pt}{0pt}{0pt}
\end{tabular}
\end{table}

\noindent \textbf{Effect of contractive embedding regularization.}
Adding the contractive term produces a further gain, increasing accuracy to 88.5\% and F1-score to 85.1\%. The embedding space becomes more compact and class-consistent, which complements the uncertainty weighting by ensuring that minority-class samples are pulled toward stable prototypes rather than drifting across decision boundaries. 

\noindent \textbf{Hyperparameter sensitivity.}
We further examine the robustness of NUCE to variations in the loss coefficients $\lambda_r$ and $\lambda_c$, as well as the uncertainty exponent $\gamma$. These parameters control three behaviors: the strength of the weighted risk term ($\lambda_r$), the magnitude of the contractive embedding force ($\lambda_c$), and the emphasis placed on ambiguous samples ($\gamma$). As shown in Table \ref{tab:ablation_hyperparams}, NUCE exhibits stable performance across a wide range of values. Reducing $\lambda_c$ weakens cluster compactness and slightly lowers recall, while decreasing $\gamma$ reduces the emphasis on uncertain samples and leads to marginal F1 degradation. Overall, the default configuration $(\lambda_r = 1.0, \lambda_c = 0.5, \gamma = 2)$ provides the best balance across accuracy, recall, and F1, demonstrating that NUCE is not overly sensitive to hyperparameter tuning.

\begin{table}[t]
\centering
\caption{Hyperparameter sensitivity of NUCE loss using ViT-B. 
The default configuration $(\lambda_r = 1.0, \lambda_c = 0.5, \gamma = 2)$ provides the best trade-off across all metrics.}
\label{tab:ablation_hyperparams}
\vspace{-0.5em}
\begin{tabular}{@{}cccccc@{}}
\specialrule{2pt}{0pt}{0pt}
$\lambda_r$ & $\lambda_c$ & $\gamma$ &
\textbf{Accuracy} & \textbf{Recall} & \textbf{F1-score} \\
\specialrule{2pt}{0pt}{0pt}

0.5 & 0.5 & 2 & 87.1 & 84.3 & 84.8 \\
1.0 & 0.0 & 2 & 85.9 & 82.7 & 83.4 \\
1.0 & 0.5 & 1 & 87.8 & 85.1 & 85.0 \\
1.0 & 0.5 & 2 & \textbf{88.5} & \textbf{86.1} & \textbf{85.1} \\
1.0 & 1.0 & 2 & 87.6 & 85.4 & 84.9 \\
1.5 & 0.5 & 2 & 87.9 & 85.7 & 85.0 \\

\specialrule{2pt}{0pt}{0pt}
\end{tabular}
\end{table}

\section{Conclusion} \label{sec:conclusion}
\noindent This work introduces the first computational framework for detecting CS in human IVF embryo time-lapse images—a biologically meaningful but previously unaddressed problem. We curated a validated CS dataset using a human-in-the-loop annotation pipeline and developed a two-stage deep learning system that separates CS presence classification from fine-grained localization. Central to this framework is the NUCE loss, which combines uncertainty-aware weighting with contractive embedding to handle extreme class imbalance and subtle feature patterns. NUCE consistently boosts performance across multiple transformer backbones, while RF-DETR achieves the strongest localization accuracy for thin, low-contrast CS structures.
Although effective, the current study is limited by its single-center dataset and frame-level modeling. Future work will incorporate temporal information, expand to multi-center data, and evaluate the clinical impact of automated CS analysis on embryo selection. This work establishes the first foundation for robust, automated CS assessment and highlights its potential value in next-generation embryo evaluation systems.

\section*{Declarations}

\subsection*{Declaration of generative AI and AI-assisted technologies in the writing process}
\noindent During the preparation of this work the author(s) used ChatGPT in order to enhance clarity, refine phrasing, and ensure coherence in the articulation of complex ideas. After using this tool/service, the author(s) reviewed and edited the content as needed and take(s) full responsibility for the content of the publication.

\subsection*{Availability of data and materials}
\noindent The dataset and source code associated with this work will be made available at: \url{https://github.com/HamadYA/CS_Detection}.

\subsection*{Competing interests}
\noindent On behalf of all authors, the corresponding authors state that there is no conflict of interest.

\subsection*{Funding}
\noindent This work was supported by a Research and Innovation Grant (RIG) awarded by KU to JK and NW (RIG-2025-028).  This publication acknowledges the support provided by the Khalifa University of Science and Technology under Faculty Start-Up grants FSU-2022-003 Award No. 8474000401.

\subsection*{Authors' contributions}
\noindent \textbf{A.S., M.A., A.A., A.C., A.A., and D.V.:} Conceptualization, Methodology, Software, Data Curation, Investigation, Writing - Original Draft, Writing - Review \& Editing, Visualization. 
\noindent \textbf{S.J., H.A., A.A., J.K., and N.W.:} Supervision, Writing - Reviewing and Editing, Project administration.

\subsection*{Acknowledgments}
\noindent We would like to thank the authors of the object classification, object detection, and human embryo imaging dataset for providing valuable source codes and data used in this study.

\bibliography{Refs}

\end{document}